\pdfoutput=1
\documentclass[runningheads]{llncs}
\usepackage{graphicx}

\usepackage{soul}
\usepackage{url}
\usepackage[utf8]{inputenc}
\usepackage{graphicx}
\usepackage{amsmath}
\usepackage{amssymb}
\usepackage{booktabs}
\usepackage{colortbl}
\usepackage{hyperref}
\usepackage{bm,array}
\hypersetup{colorlinks,urlcolor=blue,citecolor=blue,linkcolor=blue}

\begin{document}
\newcolumntype{C}{>{\centering\arraybackslash}p{2em}}
\title{Simple and Effective Graph Autoencoders with One-Hop Linear Models}

\author{
Guillaume Salha\inst{1,2}\and
Romain Hennequin\inst{1} \and
Michalis Vazirgiannis\inst{2,3}\\
}
\authorrunning{G. Salha et al.}
%
\institute{Deezer Research, Paris, France \and
LIX, \'{E}cole Polytechnique, Palaiseau, France\and
Athens University of Economics and Business, Athens, Greece\\
\email{research@deezer.com}}
\maketitle

\begin{abstract}
Over the last few years, graph autoencoders (AE) and variational autoencoders (VAE) emerged as powerful node embedding methods, with promising performances on challenging tasks such as link prediction and node clustering. Graph AE, VAE and most of their extensions rely on multi-layer graph convolutional networks (GCN) encoders to learn vector space representations of nodes. In this paper, we show that GCN encoders are actually unnecessarily complex for many applications. We propose to replace them by significantly simpler and more interpretable linear models w.r.t. the direct neighborhood (one-hop) adjacency matrix of the graph, involving fewer operations, fewer parameters and no activation function. For the two aforementioned tasks, we show that this simpler approach consistently reaches competitive performances w.r.t. GCN-based graph AE and VAE for numerous real-world graphs, including all benchmark datasets commonly used to evaluate graph AE and VAE. Based on these results, we also question the relevance of repeatedly using these datasets to compare complex graph AE and VAE. 
\end{abstract}

\section{Introduction}

Graphs have become ubiquitous, due to the proliferation of data representing relationships or interactions among entities \cite{hamilton2017representation,wu2019comprehensive}. Extracting relevant information from these entities, called the \textit{nodes} of the graph, is crucial to effectively tackle numerous machine learning tasks, such as link prediction or node clustering. While traditional approaches mainly focused on hand-engineered features \cite{bhagat2011node,liben2007link}, significant improvements were recently achieved by methods aiming at directly \textit{learning} node representations that summarize the graph structure (see \cite{hamilton2017representation} for a review). In a nutshell, these \textit{representation learning} methods aim at embedding nodes as vectors in a low-dimensional vector space in which nodes with structural proximity in the graph should be close, e.g. by leveraging random walk strategies \cite{grover2016node2vec,perozzi2014deepwalk}, matrix factorization \cite{cao2015grarep,ou2016asymmetric} or graph neural networks \cite{hamilton2017inductive,kipf2016-1}.

In particular, \textit{graph autoencoders} (AE) \cite{kipf2016-2,tian2014learning,wang2016structural} and \textit{graph variational autoencoders} (VAE) \cite{kipf2016-2} recently emerged as powerful node embedding methods. Based on encoding-decoding schemes, i.e. on the design of low dimensional vector space representations of nodes (\textit{encoding}) from which reconstructing the graph (\textit{decoding}) should be possible, graph AE and VAE models have been successfully applied to address several challenging learning tasks, with competitive results w.r.t. popular baselines such as \cite{grover2016node2vec,perozzi2014deepwalk}. These tasks include link prediction \cite{grover2019graphite,kipf2016-2,pan2018arga,salha2019-2,tran2018multi}, node clustering \cite{pan2018arga,salha2019-1,wang2017mgae}, matrix completion for inference and recommendation \cite{berg2018matrixcomp,do2019matrix} and molecular graph generation \cite{molecule3,molecule1,molecule2,simonovsky2018graphvae}. Existing models usually rely on graph neural networks (GNN) to encode nodes into embeddings. More precisely, most of them implement \textit{graph convolutional networks} (GCN) encoders \cite{do2019matrix,grover2019graphite,semiimplicit2019,huang2019rwr,kipf2016-2,pan2018arga,salha2019-1,salha2019-2,aaai20} with multiple layers, a model originally introduced by \cite{kipf2016-1}.

However, despite the prevalent use of GCN encoders in recent literature, the relevance of this design choice has never been thoroughly studied nor challenged. The actual benefit of incorporating GCNs in graph AE and VAE w.r.t. significantly simpler encoding strategies remains unclear. In this paper\footnote{A preliminary version of this work has been presented at the NeurIPS 2019 workshop on Graph Representation Learning \cite{salha2019keep}.}, we propose to tackle this important aspect, showing that GCN-based graph AE and VAE are often unnecessarily complex for numerous applications. Our work falls into a family of recent efforts questioning the systematic use of complex deep learning methods without clear comparison to less fancy but simpler baselines \cite{recsys,lin2019neural,shchur2018pitfalls}. More precisely, our contribution is threefold:
\begin{itemize}
    \item We introduce and study simpler versions of graph AE and VAE, replacing multi-layer GCN encoders by linear models w.r.t. the direct neighborhood (one-hop) adjacency matrix of the graph, involving a unique weight matrix to tune, fewer operations and no activation function.
    \item Through an extensive empirical analysis on 17 real-world graphs with various sizes and characteristics, we show that these simplified models consistently reach competitive performances w.r.t. GCN-based graph AE and VAE on link prediction and node clustering tasks. We identify the settings where simple linear encoders appear as an effective alternative to GCNs, and as first relevant baseline to implement before diving into more complex models. We also question the relevance of current benchmark datasets (Cora, Citeseer, Pubmed) commonly used in the literature to evaluate graph AE and VAE.
    \item We publicly release the code\footnote{\href{https://github.com/deezer/linear_graph_autoencoders}{https://github.com/deezer/linear\_graph\_autoencoders}} of these experiments, for reproducibility and easier future usages.
\end{itemize}

This paper is organized as follows. After reviewing key concepts on graph AE, VAE and on multi-layer GCNs in Section 2, we introduce the proposed simplified graph AE and VAE models in Section 3. We present and interpret our experiments in Section 4, and we conclude in Section 5.

\section{Preliminaries}

We consider a graph $\mathcal{G} = (\mathcal{V},\mathcal{E})$ with $|\mathcal{V}| = n$ nodes and $|\mathcal{E}| = m$ edges. In most of this paper, we assume that $\mathcal{G}$ is undirected (we extend our approach to directed graphs in Section 4.2). $A$ is the direct neighborhood (i.e. one-hop) binary adjacency matrix of $\mathcal{G}$.

\subsection{Graph Autoencoders}

Graph autoencoders (AE) \cite{kipf2016-2,tian2014learning,wang2016structural} are a family of models aiming at mapping (\textit{encoding}) each node $i \in \mathcal{V}$ of the graph $\mathcal{G}$ to a low-dimensional vector $z_i \in \mathbb{R}^d$, with $d \ll n$, from which reconstructing (\textit{decoding}) the graph should be possible. The intuition of this encoding-decoding scheme is the following: if, starting from the node embedding, the model is able to reconstruct an adjacency matrix $\hat{A}$ close to the true one, then the low-dimensional vectors $z_i$ should capture some important characteristics of the original graph structure.

Formally, the $n \times d$ matrix $Z$, whose rows are the $z_i$ vectors, is usually the output of a graph neural network (GNN) \cite{bruna2013spectral,defferrard2016,kipf2016-1} processing $A$. Then, to reconstruct the graph, most models  stack an \textit{inner product decoder} \cite{kipf2016-2} to this GNN, i.e. we have $\hat{A}_{ij} = \sigma (z^T_i z_j)$ for all node pairs $(i,j)$, with $\sigma(\cdot)$ denoting the sigmoid function: $\sigma(x) = 1/(1 + e^{-x})$. Therefore, the larger the inner product $z^T_i z_j$ in the embedding, the more likely nodes $i$ and $j$ are connected in $\mathcal{G}$ according to the AE. In a nutshell, we have:
\begin{equation}\hat{A} = \sigma(ZZ^T) \text{ with } Z = \text{GNN(A)}.
\end{equation}

Several recent works also proposed more complex decoders \cite{grover2019graphite,salha2019-2,aaai20}, that we consider as well in our experiments in Section 4. During the training phase, the GNN weights are tuned by gradient descent to iteratively minimize a \textit{reconstruction loss} capturing the similarity between $A$ and $\hat{A}$, formulated as a cross entropy loss \cite{kipf2016-2}:
\begin{equation}
\mathcal{L}^{\text{AE}} = - \frac{1}{n^2} \sum_{(i,j) \in \mathcal{V} \times \mathcal{V}} \Big[A_{ij} \log \hat{A}_{ij} + (1 - A_{ij}) \log (1 - \hat{A}_{ij}) \Big],
\end{equation}
In the above equation $(2)$, some existing methods \cite{kipf2016-2,salha2019-1,salha2019-2} also adopt a link re-weighting strategy, to reinforce the relative importance of positive links ($A_{ij} =1$) in the loss when dealing with sparse graphs.

\subsection{Graph Variational Autoencoders}

\cite{kipf2016-2} also extended the \textit{variational autoencoder} (VAE) framework from \cite{kingma2013vae} to graph structures. Authors designed a probabilistic model involving a latent variable $z_i$ of dimension $d \ll n$ for each node $i \in \mathcal{V}$, interpreted as node representations in a $d$-dimensional embedding space. In their approach, the inference model, i.e. the \textit{encoding} part of the graph VAE, is defined as:
\begin{equation}
q(Z|A) = \prod_{i=1}^n q(z_i|A) \text{ with } q(z_i|A) = \mathcal{N}(z_i|\mu_i, \text{diag}(\sigma_i^2)).\end{equation}
Gaussian means and variances parameters are learned using a GNN for each one, i.e. $\mu = \text{GNN}_{\mu}(A)$, with $\mu$ the $n \times d$ matrix stacking up $d$-dimensional mean vectors $\mu_i$ ;  likewise, $\log \sigma = \text{GNN}_{\sigma}(A)$. Latent vectors $z_i$ are samples drawn from these distributions. Then, a generative model aims at reconstructing (\textit{decoding}) $A$, leveraging inner products with sigmoid activations:
\begin{equation}
p(A|Z) = \prod_{i=1}^n \prod_{j=1}^n p(A_{ij}|z_i, z_j) \text{ with } p(A_{ij} = 1|z_i, z_j) = \hat{A}_{ij} = \sigma(z_i^Tz_j).
\end{equation}
During training, GNN weights are tuned by iteratively maximizing a tractable variational lower bound (ELBO) of the model's likelihood \cite{kipf2016-2}:
\begin{equation}
\mathcal{L}^{\text{VAE}} = \mathbb{E}_{q(Z|A)} \Big[\log
p(A|Z)\Big] - \mathcal{D}_{KL}(q(Z|A)||p(Z)),
\end{equation}
by gradient descent, with a Gaussian prior on the distribution of latent vectors, and using the \textit{reparameterization trick} from \cite{kingma2013vae}. $\mathcal{D}_{KL}(\cdot||\cdot)$ denotes the Kullback-Leibler divergence \cite{kullback1951information}.

\subsection{Graph Convolutional Networks}

While the term \textit{GNN encoder} is generic, a majority of successful applications of graph AE and VAE actually relied on multi-layer \textit{graph convolutional networks (GCN)} \cite{kipf2016-1} to encode nodes. This includes the seminal graph AE and VAE models from \cite{kipf2016-2} as well as numerous extensions \cite{do2019matrix,grover2019graphite,semiimplicit2019,huang2019rwr,pan2018arga,salha2019-1,salha2019-2,aaai20}. In a multi-layer GCN with $L$ layers ($L \geq 2$), with input layer $H^{(0)} = I_n$ and output layer $H^{(L)}$ (with $H^{(L)} = Z$ for AE, and $H^{(L)} = \mu$ or $\log \sigma$ for VAE), embedding vectors are iteratively updated, as follows:
\begin{align}
&H^{(l)} = \text{ReLU} (\tilde{A} H^{(l-1)} W^{(l-1)}), \hspace{3pt} \text{for } l \in \{1,...L-1\} \\
&H^{(L)} = \tilde{A} H^{(L-1)} W^{(L-1)} \nonumber
\end{align}
where $\tilde{A} = D^{-1/2}(A + I_n) D^{-1/2}$. $D$ is the diagonal degree matrix of $A + I_n$, and $\tilde{A}$ is therefore its symmetric normalization. At each layer, each node averages representations from its neighbors (that, from layer 2, have aggregated representations from their own neighbors), with a ReLU activation: $\text{ReLU}(x) = \max(x,0)$. $W^{(0)},...,W^{(L-1)}$ are weight matrices to tune; their dimensions can differ across layers.

GCNs became popular encoders for graph AE and VAE, thanks to their reduced computational complexity w.r.t. other GNNs \cite{bruna2013spectral,defferrard2016}, and notably the linear time complexity w.r.t. $m$ of evaluating each layer \cite{kipf2016-1}. Moreover, GCN models can also leverage node-level features, summarized in an $n \times f$ matrix $X$, in addition to the graph structure. In such setting, the input layer becomes $H^{(0)} = X$ instead of the identity matrix $I_n$.

\section{Simplifying Graph AE and VAE with One-Hop Linear Encoders}

Graph AE and VAE emerged as powerful node embedding methods with promising applications and performances \cite{berg2018matrixcomp,do2019matrix,grover2019graphite,semiimplicit2019,huang2019rwr,molecule3,molecule1,molecule2,pan2018arga,salha2019-1,salha2019-2,aaai20,simonovsky2018graphvae,wang2017mgae}. However, while almost all recent efforts from the literature implement multi-layer GCN (or an other GNN) encoders, the question of the actual benefit of such complex encoding schemes w.r.t. much simpler strategies remains widely open. In the following two sections, we tackle this important problem, arguing that these encoders often bring unnecessary complexity and redundancy. We propose and study alternative versions of graph AE and VAE, learning node embeddings from linear models, i.e. from simpler and more interpretable encoders, involving fewer parameters, fewer computations and no activation function.

\subsection{Linear Graph AE}

In this paper, we propose to replace the multi-layer GCN encoder by a simple linear model w.r.t. the normalized one-hop adjacency matrix of the graph. In the AE framework, we set:
\begin{equation}
Z = \tilde{A}  W, \text{ then } \hat{A} = \sigma(ZZ^T).
\end{equation}

We refer to this model as \textit{linear graph AE}. Embedding vectors are obtained by multiplying the $n \times n$ normalized adjacency matrix $\tilde{A}$, as defined in $(6)$, by a unique $n \times d$ weight matrix $W$. We tune this matrix in a similar fashion w.r.t. graph AE \cite{kipf2016-2}, i.e. by iteratively minimizing a weighted cross-entropy loss capturing the quality of the reconstruction $\hat{A}$ w.r.t. the original matrix $A$, by gradient descent.

This encoder is a straightforward linear mapping. Each element of $z_i$ is a weighted average from node $i$'s direct one-hop connections. Contrary to multi-layer GCN encoders (as $L\geq 2$), it ignores higher-order information ($k$-hop with $k>1$). Also, the encoder does not include any non-linear activation function. In Section 4, we will highlight the very limited impact of these two simplifications on empirical performances.

This encoder runs in a linear time w.r.t. the number of edges $m$ using a sparse representation for $\tilde{A}$, and involves fewer matrix operations than a GCN. It includes $nd$ parameters i.e. slightly fewer than the $n d + (L - 1) d^2$ parameters required by a $L$-layer GCN with $d$-dim layers. However, as for standard graph AE, the inner-product decoder has a quadratic $O(dn^2)$ complexity, as it involves the multiplication of the two dense matrices $Z$ and $Z^T$. We discuss scalability strategies in Section 4, where we also implement two alternative decoders.

Linear graph AE models can also leverage graph datasets that include node-level features vectors of dimension $f$, stacked up in an $n \times f$ matrix $X$. In such setting, the encoding step becomes:
\begin{equation}
Z = \tilde{A} X W,
\end{equation}
where the weight matrix $W$ is then of dimension $f \times d$.

\subsection{Linear Graph VAE}

We adopt a similar approach to replace the two multi-layer GCNs of standard graph VAE models by:
\begin{equation}
\mu = \tilde{A} W_{\mu} \text{ and } \log \sigma = \tilde{A} W_{\sigma},
\end{equation}
with $n \times d$ weight matrices $W_{\mu}$ and $W_{\sigma}$. Then:
\begin{equation}
\forall i \in \mathcal{V}, z_i \sim \mathcal{N}(\mu_i, \text{diag}(\sigma_i^2)),
\end{equation}
with similar decoder w.r.t. standard graph VAE (equation $(4)$). We refer to this simpler model as \textit{linear graph VAE}. During the learning phase, as standard graph VAE, we iteratively optimize the ELBO bound of equation $(5)$, w.r.t. $W_{\mu}$ and $W_{\sigma}$, by gradient descent. When dealing with graphs that include node features $X$, we instead compute:
\begin{equation}\mu = \tilde{A} X W_{\mu} \text{ and } \log \sigma = \tilde{A} X W_{\sigma},
\end{equation}
and weight matrices $W_{\mu}$ and $W_{\sigma}$ are then of dimension $f \times d$. Figure 1 displays a schematic representation of the proposed linear graph AE and VAE models.

\begin{figure}
\includegraphics[width=\textwidth]{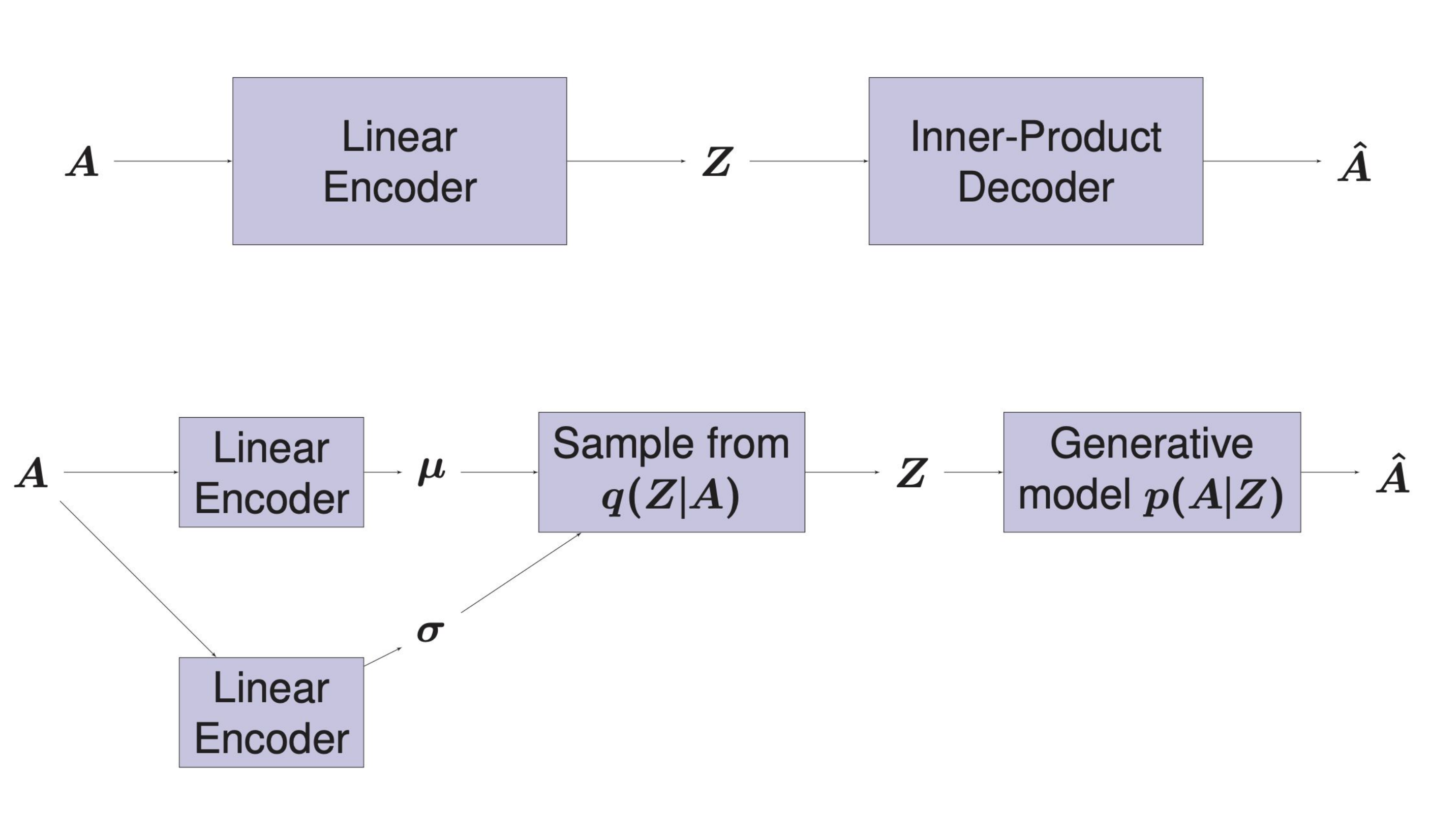}
\caption{Top: Linear Graph AE model. Bottom: Linear Graph VAE model.} \label{fig1}
\end{figure}

\subsection{Related Work}

Our work falls into a family of research efforts aiming at challenging and questioning the prevalent use of complex deep learning methods without clear comparison to simpler baselines \cite{recsys,lin2019neural,shchur2018pitfalls}. In particular, in the graph learning community, \cite{wu2019simplifying} recently proposed to simplify GCNs, notably by removing non-linearities between layers and collapsing some
weight matrices during training. Their simplified model empirically rivals standard GCNs on several large-scale classification tasks. While our work also focuses on GCNs, we argue that the two papers actually tackle very different and complementary problems:
\begin{itemize}
    \item \cite{wu2019simplifying} focus on \textit{supervised} and semi-supervised settings. They consider the GCN as the model itself, optimized to classify node-level labels. On the contrary, we consider two \textit{unsupervised} settings, in which GCNs are only a building part (the encoder) of a larger framework (the AE or the VAE), and where we optimize reconstruction losses from GCN-based embedding vectors (for AE) or from vectors drawn from distributions learned through two GCNs (for VAE).
    \item Our encoders only capture one-hop interactions: nodes aggregate information from their direct neighbors. On the contrary, \cite{wu2019simplifying} still rely on a stacked layers design that, although simplified, allows learning from higher-order interactions. Contrary to us, considering such relationships is crucial in their model for good performances (we explain in Section 4 that, in our setting, it would increase running times while bringing few to no improvement).
\end{itemize}

\section{Empirical Analysis and Discussion}

In this section, we propose an in-depth experimental evaluation of the proposed simplified graph AE and VAE models.

\subsection{Experimental Setting}

\subsubsection{Tasks.}

We consider two learning tasks. Firstly, we focus on \textit{link prediction}, as in \cite{kipf2016-2} and most subsequent works. We train models on incomplete versions of graphs where $15\%$ of edges were randomly removed. Then, we create validation and test sets from removed edges (resp. from $5\%$ and $10\%$ of edges) and from the same number of randomly sampled pairs of unconnected nodes. We evaluate the model's ability to classify edges from non-edges, using the mean \textit{Area Under the Receiver Operating Characteristic (ROC) Curve} (AUC) and \textit{Average Precision} (AP) scores on test sets, averaged over 100 runs where models were trained from 100 different random train/validation/test splits.

As a second task, we perform \textit{node clustering} from the $z_i$ vectors. When datasets include node-level ground-truth communities, we train models on complete graphs, then run $k$-means algorithms in embedding spaces. Then, we compare the resulting clusters to ground-truth communities via the mean \textit{Adjusted Mutual Information (AMI)} scores computed over 100 runs.

\subsubsection{Datasets.}

We provide experiments on 17 publicly available real-world graphs. For each graph, Tables 1 and 2 report the number of nodes $n$, the number of edges $m$, and the dimension $f$ of node features (when available):
\begin{itemize}
\item We first consider the Cora, Citeseer and Pubmed citation graphs\footnote{\label{linqs}\href{https://linqs.soe.ucsc.edu/data}{https://linqs.soe.ucsc.edu/data}}, with and without node features corresponding to $f$-dimensional bag-of-words vectors. These three graphs were used in the original experiments of \cite{kipf2016-2} and then in the wide majority of recent works \cite{grover2019graphite,semiimplicit2019,huang2019rwr,pan2018arga,park2019symmetric,salha2019-1,salha2019-2,aaai20,tran2018multi,wang2017mgae}, becoming the \textit{de facto} benchmark datasets for evaluating graph AE and VAE. Therefore, comparing linear and GCN-based models on these graphs was essential.
\item We also report results on 14 alternative graphs. We consider four other citations networks: DBLP\textsuperscript{\ref{konect}}, Arxiv-HepTh\textsuperscript{\ref{snap}}, Patent\footnote{\label{snap}\href{http://snap.stanford.edu/data/index.html}{http://snap.stanford.edu/data/index.html}} and a larger version of Cora\textsuperscript{\ref{konect}}, that we denote Cora-larger. We add the WebKD\textsuperscript{\ref{linqs}}, Blogs\textsuperscript{\ref{konect}} and Stanford\footnote{\label{konect}\href{http://konect.uni-koblenz.de/networks/}{http://konect.uni-koblenz.de/networks/}} web graphs, where hyperlinks connect web pages, as well as two Google web graphs (a medium-size one\textsuperscript{\ref{konect}}, denoted Google, and a larger one\textsuperscript{\ref{snap}}, denoted Google-large). We complete the list with two social networks (Hamsterster\textsuperscript{\ref{konect}} and LiveMocha\textsuperscript{\ref{konect}}), the Flickr\textsuperscript{\ref{konect}} image graph (nodes represent images, connected when sharing metadata), the Proteins\textsuperscript{\ref{konect}} network of proteins interactions and the Amazon\textsuperscript{\ref{konect}} products co-purchase network. We span a wide variety of real-world graphs of various natures, characteristics and sizes (from 877 to 2.7 million nodes, from 1 608 to 13.9 million edges).
\end{itemize}

\subsubsection{Models.} In all experiments, we compare the proposed simplified model to 2-layer and 3-layer GCN-based graph AE/VAE. We do not report performances of deeper models, due to significant scores deterioration. For a comparison to non-AE/VAE methods, which is out of the scope of this study, we refer to \cite{kipf2016-2,salha2019-1}.

All models were trained for $200$ epochs (resp. $300$ epochs) for graphs with $n <$ 100 000 (resp. $n \geq$ 100  000). We thoroughly checked the convergence of all models, in terms of mean AUC performances on validation sets, for these epochs numbers. As \cite{kipf2016-2}, we ignored edges directions when initial graphs were directed. For Cora, Citeseer and Pubmed, we set identical hyperparameters w.r.t. \cite{kipf2016-2} to reproduce their results, i.e. we had $d = 16$, 32-dim hidden layer(s) for GCNs, and we used Adam optimizer \cite{kingma2014adam} with a learning rate of 0.01.

For other datasets, we tuned hyperparameters by performing grid search on the validation set. We adopted a learning rate of 0.1 for Arxiv-HepTh, Patent and Stanford; of 0.05 for Amazon, Flickr, LiveMocha and Google-large; of 0.01 for Blogs, Cora-larger, DBLP, Google and Hamsterster and Proteins (AE models); of 0.005 for WebKD (except linear AE and VAE where we used 0.001 and 0.01) and Proteins (VAE models). We set $d =16$ (but we reached similar conclusions with $d =32$ and $64$), with 32-dim hidden layer(s) and without dropout.

Last, due to the prohibitive quadratic cost of reconstructing (\textit{decoding}) the exact matrix $\hat{A}$ for large graphs (with $n \geq$ 100 000), we adopted a simple stochastic sampling strategy for these graphs. At each training iteration, we estimated losses by only reconstructing a subgraph of 10 000 nodes from the original graph. These 10 000 nodes were randomly picked during training at each iteration.

\begin{table}[ht]
  \centering
\begin{scriptsize}
\caption{Link prediction on Cora, Citeseer and Pubmed benchmark datasets. Cells are grayed when linear graph AE/VAE are reaching competitive results w.r.t. standard GCN-based models (i.e. at least as good $\pm$ 1 standard deviation).}
\begin{center}
\begin{tabular}{c|cc|cc|cc}
    \toprule
    & \multicolumn{2}{c}{\textbf{Cora}} & \multicolumn{2}{c}{\textbf{Citeseer}} & \multicolumn{2}{c}{\textbf{Pubmed}} \\
     \textbf{Model} & \multicolumn{2}{c}{\tiny (n = 2 708, m = 5 429)} & \multicolumn{2}{c}{\tiny (n = 3 327, m = 4 732)} & \multicolumn{2}{c}{\tiny (n = 19 717, m = 44 338)} \\
     \cmidrule{2-7}
      & \tiny \textbf{AUC (in \%)} & \tiny \textbf{AP (in \%)} & \tiny \textbf{AUC (in \%)} & \tiny \textbf{AP (in \%)} & \tiny \textbf{AUC (in \%)} & \tiny \textbf{AP (in \%)}\\
    \midrule
    \midrule
     Linear AE (ours) & \cellcolor[gray]{.90} 83.19 $\pm$ 1.13 &  \cellcolor[gray]{.90} 87.57 $\pm$ 0.95 & \cellcolor[gray]{.90} 77.06 $\pm$ 1.81 & \cellcolor[gray]{.90} 83.05 $\pm$ 1.25 & 81.85 $\pm$ 0.32 & \cellcolor[gray]{.90} 87.54 $\pm$ 0.28 \\
     2-layer GCN AE & \cellcolor[gray]{.90} 84.79 $\pm$ 1.10 & \cellcolor[gray]{.90} 88.45 $\pm$ 0.82 & \cellcolor[gray]{.90} 78.25 $\pm$ 1.69 & \cellcolor[gray]{.90} 83.79 $\pm$ 1.24 & 82.51 $\pm$ 0.64 & \cellcolor[gray]{.90} 87.42 $\pm$ 0.38 \\
     3-layer GCN AE & \cellcolor[gray]{.90} 84.61 $\pm$ 1.22 & \cellcolor[gray]{.90} 87.65 $\pm$ 1.11 & \cellcolor[gray]{.90} 78.62 $\pm$ 1.74 & \cellcolor[gray]{.90} 82.81 $\pm$ 1.43 & 83.37 $\pm$ 0.98 & \cellcolor[gray]{.90} 87.62 $\pm$ 0.68 \\
    \midrule
    Linear VAE (ours) & \cellcolor[gray]{.90} 84.70 $\pm$ 1.24 & \cellcolor[gray]{.90} 88.24 $\pm$ 1.02 & \cellcolor[gray]{.90} 78.87 $\pm$ 1.34 & \cellcolor[gray]{.90} 83.34 $\pm$ 0.99 & \cellcolor[gray]{.90} 84.03 $\pm$ 0.28 & \cellcolor[gray]{.90} 87.98 $\pm$ 0.25 \\
    2-layer GCN VAE & \cellcolor[gray]{.90} 84.19 $\pm$ 1.07 & \cellcolor[gray]{.90} 87.68 $\pm$ 0.93 & \cellcolor[gray]{.90} 78.08 $\pm$ 1.40 & \cellcolor[gray]{.90} 83.31 $\pm$ 1.31 & \cellcolor[gray]{.90} 82.63 $\pm$ 0.45 & \cellcolor[gray]{.90} 87.45 $\pm$ 0.34 \\
     3-layer GCN VAE & \cellcolor[gray]{.90} 84.48 $\pm$ 1.42 & \cellcolor[gray]{.90} 87.61 $\pm$ 1.08 & \cellcolor[gray]{.90} 79.27 $\pm$ 1.78 & \cellcolor[gray]{.90} 83.73 $\pm$ 1.13 & \cellcolor[gray]{.90} 84.07 $\pm$ 0.47 & \cellcolor[gray]{.90} 88.18 $\pm$ 0.31 \\
    \midrule
    \midrule
       & \multicolumn{2}{c}{\textbf{Cora, with features}} & \multicolumn{2}{c}{\textbf{Citeseer, with features}} & \multicolumn{2}{c}{\textbf{Pubmed, with features}} \\
            \textbf{Model}  & \multicolumn{2}{c}{\tiny (n = 2 708, m = 5 429,} & \multicolumn{2}{c}{\tiny (n = 3 327, m = 4 732,} & \multicolumn{2}{c}{\tiny (n = 19 717, m = 44 338,} \\
            & \multicolumn{2}{c}{\tiny f = 1 433)} & \multicolumn{2}{c}{\tiny f = 3 703)} & \multicolumn{2}{c}{\tiny f = 500)} \\
     \cmidrule{2-7}
      & \tiny \textbf{AUC (in \%)} & \tiny \textbf{AP (in \%)} & \tiny \textbf{AUC (in \%)} & \tiny \textbf{AP (in \%)} & \tiny \textbf{AUC (in \%)} & \tiny\textbf{AP (in \%)}\\
    \midrule
    \midrule
    Linear AE (ours) & \cellcolor[gray]{.90} 92.05 $\pm$ 0.93 & \cellcolor[gray]{.90} 93.32 $\pm$ 0.86 & \cellcolor[gray]{.90} 91.50 $\pm$ 1.17 & \cellcolor[gray]{.90} 92.99 $\pm$ 0.97 & \cellcolor[gray]{.90} 95.88 $\pm$ 0.20 & \cellcolor[gray]{.90} 95.89 $\pm$ 0.17 \\
     2-layer GCN AE & \cellcolor[gray]{.90} 91.27 $\pm$ 0.78 & \cellcolor[gray]{.90} 92.47 $\pm$ 0.71 & \cellcolor[gray]{.90} 89.76 $\pm$ 1.39 & \cellcolor[gray]{.90} 90.32 $\pm$ 1.62 & \cellcolor[gray]{.90} 96.28 $\pm$ 0.36 & \cellcolor[gray]{.90} 96.29 $\pm$ 0.25 \\
     3-layer GCN AE & \cellcolor[gray]{.90} 89.16 $\pm$ 1.18 & \cellcolor[gray]{.90} 90.98 $\pm$ 1.01 & \cellcolor[gray]{.90} 87.31 $\pm$ 1.74 & \cellcolor[gray]{.90} 89.60 $\pm$ 1.52 & \cellcolor[gray]{.90} 94.82 $\pm$ 0.41 & \cellcolor[gray]{.90} 95.42 $\pm$ 0.26 \\
    \midrule
    Linear VAE (ours)& \cellcolor[gray]{.90} 92.55 $\pm$ 0.97 & \cellcolor[gray]{.90} 93.68 $\pm$ 0.68& \cellcolor[gray]{.90} 91.60 $\pm$ 0.90 & \cellcolor[gray]{.90} 93.08 $\pm$ 0.77 & \cellcolor[gray]{.90} 95.91 $\pm$ 0.13 & \cellcolor[gray]{.90} 95.80 $\pm$ 0.17 \\
     2-layer GCN VAE& \cellcolor[gray]{.90} 91.64 $\pm$ 0.92 & \cellcolor[gray]{.90} 92.66 $\pm$ 0.91 & \cellcolor[gray]{.90} 90.72 $\pm$ 1.01 & \cellcolor[gray]{.90} 92.05 $\pm$ 0.97 & \cellcolor[gray]{.90} 94.66 $\pm$ 0.51 & \cellcolor[gray]{.90} 94.84 $\pm$ 0.42 \\
     3-layer GCN VAE & \cellcolor[gray]{.90} 90.53 $\pm$ 0.94 & \cellcolor[gray]{.90} 91.71 $\pm$ 0.88 & \cellcolor[gray]{.90} 88.63 $\pm$ 0.95 & \cellcolor[gray]{.90} 90.20 $\pm$ 0.81 & \cellcolor[gray]{.90} 92.78 $\pm$ 1.02 & \cellcolor[gray]{.90} 93.33 $\pm$ 0.91 \\
    \bottomrule
  \end{tabular}
  \end{center}
  \end{scriptsize}
\end{table} 

\subsection{Results}

\subsubsection{Cora, Citeseer and Pubmed benchmarks.} Table 1 reports link prediction results for Cora, Citeseer and Pubmed. For standard graph AE and VAE, we managed to reproduce similar performances w.r.t. \cite{kipf2016-2}. We show that linear graph AE and VAE models consistently reach competitive performances w.r.t. 2 and 3-layer GCN-based models, i.e. they are at least as good ($\pm$ 1 standard deviation). In some settings, linear graph AE/VAE are even slightly better (e.g. $+1.25$ points in AUC for linear graph VAE on Pubmed with features, w.r.t. 2-layer GCN-based graph VAE). These results emphasize the effectiveness of the proposed simple encoding scheme on these datasets, where the empirical benefit of multi-layer GCNs is very limited. In Table 3, we consolidate our results by reaching similar conclusions on the node clustering task. Nodes are documents clustered in respectively 6, 7 and 3 topic classes, acting as ground-truth communities. In almost all settings, linear graph AE and VAE rival their GCN-based counterparts (e.g. $+4.31$ MI points for linear graph VAE on Pubmed with features, w.r.t. 2-layer GCN-based graph VAE).
\begin{table}[!ht]
  \centering
\caption{Link prediction on alternative real-world datasets. Cells are grayed when linear graph AE/VAE are reaching competitive results w.r.t. standard GCN-based models (i.e. at least as good $\pm$ 1 standard deviation).}
\begin{scriptsize}
  \begin{tabular}{c|cc|cc|cc}
    \toprule
   & \multicolumn{2}{c}{\textbf{WebKD}} & \multicolumn{2}{c}{\textbf{WebKD, with features}} & \multicolumn{2}{c}{\textbf{Hamsterster}} \\
       \textbf{Model}  & \multicolumn{2}{c}{\tiny (n = 877, m = 1 608)} & \multicolumn{2}{c}{\tiny (n = 877, m = 1 608, f = 1 703)} & \multicolumn{2}{c}{\tiny (n = 1 858, m = 12 534)}  \\
     \cmidrule{2-7}
       & \tiny \textbf{AUC (in \%)} & \tiny \textbf{AP (in \%)} & \tiny \textbf{AUC (in \%)} & \tiny \textbf{AP (in \%)} & \tiny \textbf{AUC (in \%)} & \tiny \textbf{AP (in \%)} \\
    \midrule
    \midrule
    Linear AE (ours) & \cellcolor[gray]{.90} 77.20 $\pm$ 2.35 & \cellcolor[gray]{.90} 83.55 $\pm$ 1.81 & \cellcolor[gray]{.90} 84.15 $\pm$ 1.64 & \cellcolor[gray]{.90} 87.01 $\pm$ 1.48 & \cellcolor[gray]{.90} 93.07 $\pm$ 0.67 & \cellcolor[gray]{.90} 94.20 $\pm$ 0.58 \\
     2-layer GCN AE & \cellcolor[gray]{.90} 77.88 $\pm$ 2.57 & \cellcolor[gray]{.90} 84.12 $\pm$ 2.18 & \cellcolor[gray]{.90} 86.03 $\pm$ 3.97 & \cellcolor[gray]{.90} 87.97 $\pm$ 2.76 & \cellcolor[gray]{.90} 92.07 $\pm$ 0.63 & \cellcolor[gray]{.90} 93.01 $\pm$ 0.69 \\
     3-layer GCN AE & \cellcolor[gray]{.90} 78.20 $\pm$ 3.69 & \cellcolor[gray]{.90} 83.13 $\pm$ 2.58 & \cellcolor[gray]{.90} 81.39 $\pm$ 3.93 & \cellcolor[gray]{.90} 85.34 $\pm$ 2.92 & \cellcolor[gray]{.90} 91.40 $\pm$ 0.79 & \cellcolor[gray]{.90} 92.22 $\pm$ 0.85 \\
    \midrule
    Linear VAE (ours)& \cellcolor[gray]{.90} 83.50 $\pm$ 1.98 & \cellcolor[gray]{.90} 86.70 $\pm$ 1.53 & \cellcolor[gray]{1.0} 85.57 $\pm$ 2.18 & \cellcolor[gray]{.90} 88.08 $\pm$ 1.76 & \cellcolor[gray]{.90} 91.08 $\pm$ 0.70 & \cellcolor[gray]{.90} 91.85 $\pm$ 0.64 \\
     2-layer GCN VAE & \cellcolor[gray]{.90} 82.31 $\pm$ 2.55 & \cellcolor[gray]{.90} 86.15 $\pm$ 2.03 & \cellcolor[gray]{1.0} 87.87 $\pm$ 2.48 & \cellcolor[gray]{.90} 88.97 $\pm$ 2.17 & \cellcolor[gray]{.90} 91.62 $\pm$ 0.60 & \cellcolor[gray]{.90} 92.43 $\pm$ 0.64 \\
     3-layer GCN VAE& \cellcolor[gray]{.90} 82.17 $\pm$ 2.70 & \cellcolor[gray]{.90} 85.35 $\pm$ 2.25 & \cellcolor[gray]{1.0} 89.69 $\pm$ 1.80 & \cellcolor[gray]{.90} 89.90 $\pm$ 1.58 & \cellcolor[gray]{.90} 91.06 $\pm$ 0.71 & \cellcolor[gray]{.90} 91.85 $\pm$ 0.77 \\
    \midrule
    \midrule
    & \multicolumn{2}{c}{\textbf{DBLP}} & \multicolumn{2}{c}{\textbf{Cora-larger}} & \multicolumn{2}{c}{\textbf{Arxiv-HepTh}}  \\
        \textbf{Model} & \multicolumn{2}{c}{\tiny (n = 12 591, m = 49 743)} & \multicolumn{2}{c}{\tiny (n = 23 166, m = 91 500)} & \multicolumn{2}{c}{\tiny (n = 27 770, m = 352 807)}\\
     \cmidrule{2-7}
      & \tiny \textbf{AUC (in \%)} & \tiny \textbf{AP (in \%)} & \tiny \textbf{AUC (in \%)} & \tiny \textbf{AP (in \%)} & \tiny \textbf{AUC (in \%)} & \tiny \textbf{AP (in \%)} \\
    \midrule
    \midrule
    Linear AE (ours) & \cellcolor[gray]{.90} 90.11 $\pm$ 0.40 & \cellcolor[gray]{.90} 93.15 $\pm$ 0.28 & \cellcolor[gray]{.90} 94.64 $\pm$ 0.08 & \cellcolor[gray]{.90} 95.96 $\pm$ 0.10 & \cellcolor[gray]{.90} 98.34 $\pm$ 0.03 & \cellcolor[gray]{.90} 98.46 $\pm$ 0.03 \\
     2-layer GCN AE & \cellcolor[gray]{.90} 90.29 $\pm$ 0.39 & \cellcolor[gray]{.90} 93.01 $\pm$ 0.33 & \cellcolor[gray]{.90} 94.80 $\pm$ 0.08 & \cellcolor[gray]{.90} 95.72 $\pm$ 0.05 & \cellcolor[gray]{.90} 97.97 $\pm$ 0.09 & \cellcolor[gray]{.90} 98.12 $\pm$ 0.09 \\
     3-layer GCN AE & \cellcolor[gray]{.90} 89.91 $\pm$ 0.61 & \cellcolor[gray]{.90} 92.24 $\pm$ 0.67 & \cellcolor[gray]{.90} 94.51 $\pm$ 0.31 & \cellcolor[gray]{.90} 95.11 $\pm$ 0.28 & \cellcolor[gray]{.90} 94.35 $\pm$ 1.30 & \cellcolor[gray]{.90} 94.46 $\pm$ 1.31\\
    \midrule
    Linear VAE (ours) & \cellcolor[gray]{.90} 90.62 $\pm$ 0.30 & \cellcolor[gray]{.90} 93.25 $\pm$ 0.22 & \cellcolor[gray]{.90} 95.20 $\pm$ 0.16 & \cellcolor[gray]{.90} 95.99 $\pm$ 0.12 & \cellcolor[gray]{.90} 98.35 $\pm$ 0.05 & \cellcolor[gray]{.90} 98.46 $\pm$ 0.05\\
     2-layer GCN VAE & \cellcolor[gray]{.90} 90.40 $\pm$ 0.43 & \cellcolor[gray]{.90} 93.09 $\pm$ 0.35 & \cellcolor[gray]{.90} 94.60 $\pm$ 0.20 & \cellcolor[gray]{.90} 95.74 $\pm$ 0.13 & \cellcolor[gray]{.90} 97.75 $\pm$ 0.08 & \cellcolor[gray]{.90} 97.91 $\pm$ 0.06 \\
     3-layer GCN VAE & \cellcolor[gray]{.90} 89.92 $\pm$ 0.59 & \cellcolor[gray]{.90} 92.52 $\pm$ 0.48 & \cellcolor[gray]{.90} 94.48 $\pm$ 0.28 & \cellcolor[gray]{.90} 95.30 $\pm$ 0.22 & \cellcolor[gray]{.90} 94.57 $\pm$ 1.14 & \cellcolor[gray]{.90} 94.73 $\pm$ 1.12 \\
    \midrule
    \midrule
    & \multicolumn{2}{c}{\textbf{LiveMocha}} & \multicolumn{2}{c}{\textbf{Flickr}} & \multicolumn{2}{c}{\textbf{Patent}} \\
        \textbf{Model}  & \multicolumn{2}{c}{\tiny (n = 104 103, m = 2 193 083)} & \multicolumn{2}{c}{\tiny (n = 105 938, m = 2 316 948)} &\multicolumn{2}{c}{\tiny (n = 2 745 762, m = 13 965 410)} \\
     \cmidrule{2-7} 
     & \tiny \textbf{AUC (in \%)} & \tiny \textbf{AP (in \%)} & \tiny \textbf{AUC (in \%)} & \tiny \textbf{AP (in \%)} & \tiny \textbf{AUC (in \%)} & \tiny \textbf{AP (in \%)} \\
    \midrule
    \midrule
    Linear AE (ours) &\cellcolor[gray]{.90} 93.35 $\pm$ 0.10 & \cellcolor[gray]{.90} 94.83 $\pm$ 0.08 & \cellcolor[gray]{.90} 96.38 $\pm$ 0.05 & \cellcolor[gray]{.90} 97.27 $\pm$ 0.04 & \cellcolor[gray]{.90} 85.49 $\pm$ 0.09 & \cellcolor[gray]{.90} 87.17 $\pm$ 0.07\\
     2-layer GCN AE & \cellcolor[gray]{.90} 92.79 $\pm$ 0.17 & \cellcolor[gray]{.90} 94.33 $\pm$ 0.13 & \cellcolor[gray]{.90} 96.34 $\pm$ 0.05 & \cellcolor[gray]{.90} 97.22 $\pm$ 0.04 & \cellcolor[gray]{.90} 82.86 $\pm$ 0.20 & \cellcolor[gray]{.90} 84.52 $\pm$ 0.24\\
     3-layer GCN AE & \cellcolor[gray]{.90} 92.22 $\pm$ 0.73 & \cellcolor[gray]{.90} 93.67 $\pm$ 0.57 & \cellcolor[gray]{.90} 96.06 $\pm$ 0.08 & \cellcolor[gray]{.90} 97.01 $\pm$ 0.05 & \cellcolor[gray]{.90} 83.77 $\pm$ 0.41 & \cellcolor[gray]{.90} 84.73 $\pm$ 0.42  \\
    \midrule
    Linear VAE (ours) & \cellcolor[gray]{.90} 93.23 $\pm$ 0.06 & \cellcolor[gray]{.90} 94.61 $\pm$ 0.05 & \cellcolor[gray]{1.0} 96.05 $\pm$ 0.08 & \cellcolor[gray]{0.90} 97.12 $\pm$ 0.06 & \cellcolor[gray]{0.90} 84.57 $\pm$ 0.27 & \cellcolor[gray]{0.90} 85.46 $\pm$ 0.30\\
     2-layer GCN VAE & \cellcolor[gray]{.90} 92.68 $\pm$ 0.21 & \cellcolor[gray]{.90} 94.23 $\pm$ 0.15 & \cellcolor[gray]{1.0} 96.35 $\pm$ 0.07 & \cellcolor[gray]{.90} 97.20 $\pm$ 0.06 & \cellcolor[gray]{0.90} 83.77 $\pm$ 0.28 & \cellcolor[gray]{0.90} 83.37 $\pm$ 0.26  \\
     3-layer GCN VAE & \cellcolor[gray]{.90} 92.71 $\pm$ 0.37 & \cellcolor[gray]{.90} 94.01 $\pm$ 0.26 & \cellcolor[gray]{1.0} 96.39 $\pm$ 0.13 & \cellcolor[gray]{0.90} 97.16 $\pm$ 0.08 & \cellcolor[gray]{0.90} 85.30 $\pm$ 0.51 & \cellcolor[gray]{0.90} 86.14 $\pm$ 0.49  \\
    \midrule
    \midrule
    & \multicolumn{2}{c}{\textbf{Blogs}} & \multicolumn{2}{c}{\textbf{Amazon}} & \multicolumn{2}{c}{\textbf{Google-large}} \\
        \textbf{Model} & \multicolumn{2}{c}{\tiny (n = 1 224, m = 19 025)}  & \multicolumn{2}{c}{\tiny (n = 334 863, m = 925 872)} & \multicolumn{2}{c}{\tiny (n = 875 713, m = 5 105 039)} \\
     \cmidrule{2-7} 
     & \tiny \textbf{AUC (in \%)} & \tiny \textbf{AP (in \%)} & \tiny \textbf{AUC (in \%)} & \tiny \textbf{AP (in \%)} & \tiny \textbf{AUC (in \%)} & \tiny \textbf{AP (in \%)} \\
    \midrule
    \midrule
    Linear AE (ours) & \cellcolor[gray]{.90} 91.71 $\pm$ 0.39 & \cellcolor[gray]{.90} 92.53 $\pm$ 0.44 & \cellcolor[gray]{.90} 90.70 $\pm$ 0.09 & \cellcolor[gray]{.90} 93.46 $\pm$ 0.08  & \cellcolor[gray]{.90} 95.37 $\pm$ 0.05 & \cellcolor[gray]{.90} 96.93 $\pm$ 0.05 \\
     2-layer GCN AE & \cellcolor[gray]{.90} 91.57 $\pm$ 0.34 & \cellcolor[gray]{.90} 92.51 $\pm$ 0.29 & \cellcolor[gray]{.90} 90.15 $\pm$ 0.15 & \cellcolor[gray]{.90} 92.33 $\pm$ 0.14 & \cellcolor[gray]{.90} 95.06 $\pm$ 0.08 & \cellcolor[gray]{.90} 96.40 $\pm$ 0.07 \\
     3-layer GCN AE & \cellcolor[gray]{.90} 91.74 $\pm$ 0.37 & \cellcolor[gray]{.90} 92.62 $\pm$ 0.31  & \cellcolor[gray]{.90} 88.54 $\pm$ 0.37 & \cellcolor[gray]{.90} 90.47 $\pm$ 0.38 & \cellcolor[gray]{.90} 93.68 $\pm$ 0.15 & \cellcolor[gray]{.90} 94.99 $\pm$ 0.14  \\
    \midrule
    Linear VAE (ours) & 91.34 $\pm$ 0.24 & 92.10 $\pm$ 0.24 & \cellcolor[gray]{1.0} 84.53 $\pm$ 0.08 & \cellcolor[gray]{1.0} 87.79 $\pm$ 0.06 & 91.13 $\pm$ 0.14 & 93.79 $\pm$ 0.10 \\
     2-layer GCN VAE & 91.85 $\pm$ 0.22 & 92.60 $\pm$ 0.25 & \cellcolor[gray]{1.0} 90.14 $\pm$ 0.22 & \cellcolor[gray]{1.0} 92.33 $\pm$ 0.23 & 95.04 $\pm$ 0.09 & 96.38 $\pm$ 0.07 \\
     3-layer GCN VAE & 91.83 $\pm$ 0.48 & 92.65 $\pm$ 0.35 & \cellcolor[gray]{1.0} 89.44 $\pm$ 0.25 & \cellcolor[gray]{1.0} 91.23 $\pm$ 0.23 & 93.79 $\pm$ 0.22 & 95.12 $\pm$ 0.21 \\
    \midrule
    \midrule
    & \multicolumn{2}{c}{\textbf{Stanford}} & \multicolumn{2}{c}{\textbf{Proteins}} & \multicolumn{2}{c}{\textbf{Google}} \\
        \textbf{Model}  & \multicolumn{2}{c}{\tiny (n = 281 903, m = 2 312 497)} & \multicolumn{2}{c}{\tiny (n = 6 327, m = 147 547)} & \multicolumn{2}{c}{\tiny (n = 15 763, m = 171 206)} \\
     \cmidrule{2-7} 
     & \tiny \textbf{AUC (in \%)} & \tiny \textbf{AP (in \%)} & \tiny \textbf{AUC (in \%)} & \tiny \textbf{AP (in \%)} & \tiny \textbf{AUC (in \%)} & \tiny \textbf{AP (in \%)} \\
    \midrule
    \midrule
    Linear AE (ours) & \cellcolor[gray]{.90} 97.73 $\pm$ 0.10 & \cellcolor[gray]{.90} 98.37 $\pm$ 0.10 & \cellcolor[gray]{1.0} 94.09 $\pm$ 0.23 & \cellcolor[gray]{1.0} 96.01 $\pm$ 0.16 & \cellcolor[gray]{1.0} 96.02 $\pm$ 0.14 & \cellcolor[gray]{1.0} 97.09 $\pm$ 0.08 \\
     2-layer GCN AE & \cellcolor[gray]{.90} 97.05 $\pm$ 0.63 & \cellcolor[gray]{.90} 97.56 $\pm$ 0.55  & \cellcolor[gray]{1.0} 94.55 $\pm$ 0.20 & \cellcolor[gray]{1.0} 96.39 $\pm$ 0.16 & \cellcolor[gray]{1.0} 96.66 $\pm$ 0.24 & \cellcolor[gray]{1.0} 97.45 $\pm$ 0.25 \\
     3-layer GCN AE & \cellcolor[gray]{.90} 92.19 $\pm$ 1.49 & \cellcolor[gray]{.90} 92.58 $\pm$ 1.50 & \cellcolor[gray]{1.0} 94.30 $\pm$ 0.19 & \cellcolor[gray]{1.0} 96.08 $\pm$ 0.15 & \cellcolor[gray]{1.0} 95.10 $\pm$ 0.27 & \cellcolor[gray]{1.0} 95.94 $\pm$ 0.20  \\
    \midrule
    Linear VAE (ours) & \cellcolor[gray]{1.0} 94.96 $\pm$ 0.25 & \cellcolor[gray]{1.0} 96.64 $\pm$ 0.15 & 93.99 $\pm$ 0.10 & \cellcolor[gray]{0.90} 95.94 $\pm$ 0.16 & 91.11 $\pm$ 0.31 & 92.91 $\pm$ 0.18\\
     2-layer GCN VAE & \cellcolor[gray]{1.0} 97.60 $\pm$ 0.11 & \cellcolor[gray]{1.0} 98.02 $\pm$ 0.10 & 94.57 $\pm$ 0.18 & \cellcolor[gray]{0.90} 96.18 $\pm$ 0.33 & 96.11 $\pm$ 0.59 & 96.84 $\pm$ 0.51 \\
     3-layer GCN VAE &\cellcolor[gray]{1.0} 97.53 $\pm$ 0.13 & \cellcolor[gray]{1.0} 98.01 $\pm$ 0.10& 94.27 $\pm$ 0.25 & \cellcolor[gray]{0.90} 95.71 $\pm$ 0.28 & 95.10 $\pm$ 0.54 & 96.00 $\pm$ 0.44 \\
    \bottomrule
  \end{tabular}
  \end{scriptsize}
\end{table} 

\subsubsection{Alternative graph datasets.}
Table 2 reports link prediction results for all other graphs. Linear graph AE models are competitive in 13 cases out of 15, and sometimes even achieve better performances (e.g. $+1.72$ AUC points for linear graph AE on the largest dataset, Patent, w.r.t. 3-layer GCN-based graph AE). Moreover, linear graph VAE models rival or outperform GCN-based models in 10 cases out of 15. Overall, linear graph AE/VAE also achieve very close results w.r.t. GCN-based models in all remaining datasets (e.g. on Google, with a mean AUC score of 96.02\% $\pm$ 0.14 for linear graph AE, only slightly below the mean AUC score of 96.66\% $\pm$ 0.24 of 2-layer GCN-based graph AE). This confirms the empirical effectiveness of simple node encoding schemes, that appear as a suitable alternative to complex multi-layer encoders for many real-world applications. Regarding node clustering (Table 3), linear AE and VAE models are competitive on the Cora-larger graph, in which nodes are documents clustered in 70 topic classes. However, 2-layer and 3-layer GCN-based models are significantly outperforming on the Blogs graph, where political blogs are classified as either left-leaning or right-leaning (e.g. $-23.42$ MI points for linear graph VAE w.r.t. 2-layer GCN-based graph VAE).

\begin{table}[t]
  \centering
\begin{scriptsize}
\caption{Node clustering on graphs with communities. Cells are grayed when linear graph AE/VAE are reaching competitive results w.r.t. standard GCN-based models (i.e. at least as good $\pm$ 1 standard deviation).}
  \begin{tabular}{c|c|c|c|c}
    \toprule
    & \textbf{Cora} & \textbf{Cora with features} & \textbf{Citeseer} & \textbf{Citeseer with features}\\
      \textbf{Model}  & \tiny (n = 2 708, & \tiny (n = 2 708,  m = 5 429, & \tiny (n = 3 327, & \tiny (n = 3 327, m = 4 732,\\
      &  \tiny m = 5 429) & \tiny f = 1 433) & \tiny m = 4 732) & \tiny f = 3 703)\\
     \cmidrule{2-5}
      & \tiny \textbf{AMI (in \%)} & \tiny \textbf{AMI (in \%)} & \tiny \textbf{AMI (in \%)} & \tiny \textbf{AMI (in \%)}\\
    \midrule
    \midrule
    Linear AE (ours) & 26.31 $\pm$ 2.85 & \cellcolor[gray]{.90} 47.02 $\pm$ 2.09 & \cellcolor[gray]{.90} 8.56 $\pm$ 1.28 & \cellcolor[gray]{.90} 20.23 $\pm$ 1.36 \\
     2-layer GCN AE & 30.88 $\pm$ 2.56 & \cellcolor[gray]{.90} 43.04 $\pm$ 3.28 & \cellcolor[gray]{.90} 9.46 $\pm$ 1.06 & \cellcolor[gray]{.90} 19.38 $\pm$ 3.15 \\
     3-layer GCN AE & 33.06 $\pm$ 3.10 & \cellcolor[gray]{.90} 44.12 $\pm$ 2.48 & \cellcolor[gray]{.90} 10.69 $\pm$ 1.98 & \cellcolor[gray]{.90} 19.71 $\pm$ 2.55 \\
     \midrule
         Linear VAE (ours) & \cellcolor[gray]{.90} 34.35 $\pm$ 1.42 & \cellcolor[gray]{.90} 48.12 $\pm$ 1.96 & \cellcolor[gray]{.90} 12.67 $\pm$ 1.27 & \cellcolor[gray]{.90} 20.71 $\pm$ 1.95  \\
     2-layer GCN VAE & \cellcolor[gray]{.90} 26.66 $\pm$ 3.94 & \cellcolor[gray]{.90} 44.84 $\pm$ 2.63 & \cellcolor[gray]{.90} 9.85 $\pm$ 1.24 & \cellcolor[gray]{.90} 20.17 $\pm$ 3.07 \\
     3-layer GCN VAE & \cellcolor[gray]{.90} 28.43 $\pm$ 2.83 & \cellcolor[gray]{.90} 44.29 $\pm$ 2.54 & \cellcolor[gray]{.90} 10.64 $\pm$ 1.47 & \cellcolor[gray]{.90} 19.94 $\pm$ 2.50  \\
    \midrule
    \midrule
    & \textbf{Pubmed} & \textbf{Pubmed with features} & \textbf{Cora-larger} & \textbf{Blogs}\\
       \textbf{Model}  & \tiny (n = 19 717, & \tiny (n = 19 717, m = 44 338,  & \tiny (n = 23 166, & \tiny (n = 1 224,  m = 19 025)\\
     &  \tiny m = 44 338) & \tiny f = 500) & \tiny  m = 91 500) & \\
     \cmidrule{2-5}
      & \tiny \textbf{AMI (in \%)} & \tiny \textbf{AMI (in \%)} & \tiny \textbf{AMI (in \%)} & \tiny \textbf{AMI (in \%)}\\
    \midrule
    \midrule
    Linear AE (ours) & 10.76 $\pm$ 3.70 & \cellcolor[gray]{.90} 26.12 $\pm$ 1.94 & \cellcolor[gray]{.90} 40.34 $\pm$ 0.51 & 46.84 $\pm$ 1.79 \\
     2-layer GCN AE & 16.41 $\pm$ 3.15 & \cellcolor[gray]{.90} 23.08 $\pm$ 3.35 & \cellcolor[gray]{.90} 39.75 $\pm$ 0.79 & 72.58 $\pm$ 4.54 \\
     3-layer GCN AE & 23.11 $\pm$ 2.58 & \cellcolor[gray]{.90} 25.94 $\pm$ 3.09 & \cellcolor[gray]{.90} 35.67 $\pm$ 1.76 & 72.72 $\pm$ 1.80 \\
     \midrule
         Linear VAE (ours) & \cellcolor[gray]{.90} 25.14 $\pm$ 2.83 & \cellcolor[gray]{.90} 29.74 $\pm$ 0.64 & \cellcolor[gray]{.90} 43.32 $\pm$ 0.52 & \cellcolor[gray]{1.0} 49.70 $\pm$ 1.08 \\
     2-layer GCN VAE & \cellcolor[gray]{.90} 20.52 $\pm$ 2.97 & \cellcolor[gray]{.90} 25.43 $\pm$ 1.47 & \cellcolor[gray]{.90} 38.34 $\pm$ 0.64 & \cellcolor[gray]{1.0} 73.12 $\pm$ 0.83 \\
     3-layer GCN VAE & \cellcolor[gray]{.90} 21.32 $\pm$ 3.70 & \cellcolor[gray]{.90} 24.91 $\pm$ 3.09 & \cellcolor[gray]{.90} 37.30 $\pm$ 1.07 & \cellcolor[gray]{1.0} 70.56 $\pm$ 5.43 \\
    \bottomrule
  \end{tabular}
  \end{scriptsize}
\end{table}

\subsubsection{Experiments on more complex decoders.}
So far, we compared different encoders but the (standard) inner-product decoder was fixed. As a robustness check, in Table 4, we report complementary link prediction experiments, on variants of graph AE/VAE with two more complex decoders from recent works:
\begin{itemize}
    \item The Graphite model from \cite{grover2019graphite}, that still considers undirected graphs, but rely on an iterative graph refinement strategy inspired by low-rank
approximations for decoding.
\item The Gravity-Inspired model from \cite{salha2019-2}, that provides an asymmetric decoding scheme (i.e. $\hat{A}_{ij} \neq \hat{A}_{ji}$). This model handles directed graphs. Therefore, contrary to previous experiments, we do not ignore edges directions when initial graphs were directed.
\end{itemize}

We draw similar conclusions w.r.t. Tables 1 and 2, consolidating our conclusions. For brevity, we only report results for the Cora, Citeseer and Pubmed graphs, where linear models are competitive, and for the Google graph, where GCN-based graph AE and VAE slightly outperform. We stress out that scores from Graphite \cite{grover2019graphite} and Gravity \cite{salha2019-2} models are \textit{not} directly comparable, as the former ignores edges directionalities while the latter processes directed graphs, i.e. the learning task becomes a \textit{directed} link prediction problem.

\begin{table}[t]
  \centering
\begin{scriptsize}
\caption{Link prediction with Graphite and Gravity alternative decoding schemes. Cells are grayed when linear graph AE/VAE are reaching competitive results w.r.t. GCN-based models (i.e. at least as good $\pm$ 1 standard deviation).}
  \begin{tabular}{c|cc|cc}
    \toprule
    & \multicolumn{2}{c}{\textbf{Cora}} & \multicolumn{2}{c}{\textbf{Citeseer}} \\
     \textbf{Model} & \multicolumn{2}{c}{\tiny (n = 2 708, m = 5 429)} & \multicolumn{2}{c}{\tiny (n = 3 327, m = 4 732)} \\
     \cmidrule{2-5}
      & \tiny \textbf{AUC (in \%)} & \tiny \textbf{AP (in \%)} & \tiny \textbf{AUC (in \%)} & \tiny \textbf{AP (in \%)}\\
    \midrule
    \midrule
    Linear Graphite AE (ours) & \cellcolor[gray]{.90} 83.42 $\pm$ 1.76 & \cellcolor[gray]{.90} 87.32 $\pm$ 1.53 & \cellcolor[gray]{.90} 77.56 $\pm$ 1.41 & \cellcolor[gray]{.90} 82.88 $\pm$ 1.15 \\
     2-layer Graphite AE & \cellcolor[gray]{.90} 81.20 $\pm$ 2.21 & \cellcolor[gray]{.90} 85.11 $\pm$ 1.91 & \cellcolor[gray]{.90} 73.80 $\pm$ 2.24 & \cellcolor[gray]{.90} 79.32 $\pm$ 1.83 \\
     3-layer Graphite AE & \cellcolor[gray]{.90} 79.06 $\pm$ 1.70 & \cellcolor[gray]{.90} 81.79 $\pm$ 1.62 & \cellcolor[gray]{.90} 72.24 $\pm$ 2.29 & \cellcolor[gray]{.90} 76.60 $\pm$ 1.95 \\
    \midrule
    Linear Graphite VAE (ours) & \cellcolor[gray]{.90} 83.68 $\pm$ 1.42 & \cellcolor[gray]{.90} 87.57 $\pm$ 1.16 & \cellcolor[gray]{.90} 78.90 $\pm$ 1.08 & \cellcolor[gray]{.90} 83.51 $\pm$ 0.89 \\
     2-layer Graphite VAE  & \cellcolor[gray]{.90} 84.89 $\pm$ 1.48 & \cellcolor[gray]{.90} 88.10 $\pm$ 1.22 & \cellcolor[gray]{.90} 77.92 $\pm$ 1.57 & \cellcolor[gray]{.90} 82.56 $\pm$ 1.31 \\
     3-layer Graphite VAE  & \cellcolor[gray]{.90} 85.33 $\pm$ 1.19 & \cellcolor[gray]{.90} 87.98 $\pm$ 1.09 & \cellcolor[gray]{.90} 77.46 $\pm$ 2.34 & \cellcolor[gray]{.90} 81.95 $\pm$ 1.71\\
    \midrule
    \midrule
    Linear Gravity AE (ours) & \cellcolor[gray]{.90} 90.71 $\pm$ 0.95 & \cellcolor[gray]{.90} 92.95 $\pm$ 0.88 & \cellcolor[gray]{.90} 80.52 $\pm$ 1.37 & \cellcolor[gray]{.90} 86.29 $\pm$ 1.03 \\
     2-layer Gravity AE & \cellcolor[gray]{.90} 87.79 $\pm$ 1.07 & \cellcolor[gray]{.90} 90.78 $\pm$ 0.82 & \cellcolor[gray]{.90} 78.36 $\pm$ 1.55 & \cellcolor[gray]{.90} 84.75 $\pm$ 1.10 \\
     3-layer Gravity AE & \cellcolor[gray]{.90} 87.76 $\pm$ 1.32 & \cellcolor[gray]{.90} 90.15 $\pm$ 1.45 & \cellcolor[gray]{.90} 78.32 $\pm$ 1.92 & \cellcolor[gray]{.90} 84.88 $\pm$ 1.36 \\
    \midrule
    Linear Gravity VAE (ours) & \cellcolor[gray]{.90} 91.29 $\pm$ 0.70 & \cellcolor[gray]{.90} 93.01 $\pm$ 0.57 & \cellcolor[gray]{.90} 86.65 $\pm$ 0.95 & \cellcolor[gray]{.90} 89.49 $\pm$ 0.69 \\
     2-layer Gravity VAE  & \cellcolor[gray]{.90} 91.92 $\pm$ 0.75 & \cellcolor[gray]{.90} 92.46 $\pm$ 0.64 & \cellcolor[gray]{.90} 87.67 $\pm$ 1.07 & \cellcolor[gray]{.90} 89.79 $\pm$ 1.01\\
     3-layer Gravity VAE  & \cellcolor[gray]{.90} 90.80 $\pm$ 1.28 & \cellcolor[gray]{.90} 92.01 $\pm$ 1.19 & \cellcolor[gray]{.90} 85.28 $\pm$ 1.33 & \cellcolor[gray]{.90} 87.54 $\pm$ 1.21 \\
    \midrule
    \midrule
    & \multicolumn{2}{c}{\textbf{Pubmed}} & \multicolumn{2}{c}{\textbf{Google}} \\
     \textbf{Model} & \multicolumn{2}{c}{\tiny (n = 19 717, m = 44 338)} & \multicolumn{2}{c}{\tiny (n = 15 763, m = 171 206)} \\
     \cmidrule{2-5}
      & \tiny \textbf{AUC (in \%)} & \tiny \textbf{AP (in \%)} & \tiny \textbf{AUC (in \%)} & \tiny \textbf{AP (in \%)}\\
    \midrule
    \midrule
    Linear Graphite AE (ours) & \cellcolor[gray]{.90} 80.28 $\pm$ 0.86 & \cellcolor[gray]{.90} 85.81 $\pm$ 0.67 & 94.30 $\pm$ 0.22  & 95.09 $\pm$ 0.16 \\
     2-layer Graphite AE & \cellcolor[gray]{.90} 79.98 $\pm$ 0.66 & \cellcolor[gray]{.90} 85.33 $\pm$ 0.41 & 95.54 $\pm$ 0.42  & 95.99 $\pm$ 0.39 \\
     3-layer Graphite AE & \cellcolor[gray]{.90} 79.96 $\pm$ 1.40 & \cellcolor[gray]{.90} 84.88 $\pm$ 0.89 & 93.99 $\pm$ 0.54 & 94.74 $\pm$ 0.49 \\
    \midrule
    Linear Graphite VAE (ours) & 79.59 $\pm$ 0.33 & 86.17 $\pm$ 0.31 & 92.71 $\pm$ 0.38  & 94.41 $\pm$ 0.25 \\
     2-layer Graphite VAE  & 82.74 $\pm$ 0.30 & 87.19 $\pm$ 0.36 & 96.49 $\pm$ 0.22  & 96.91 $\pm$ 0.17 \\
     3-layer Graphite VAE  & 84.56 $\pm$ 0.42 & 88.01 $\pm$ 0.39 & 96.32 $\pm$ 0.24  & 96.62 $\pm$ 0.20 \\
    \midrule
    \midrule
    Linear Gravity AE (ours) & \cellcolor[gray]{.90} 76.78 $\pm$ 0.38 & \cellcolor[gray]{.90} 84.50 $\pm$ 0.32 & 97.46 $\pm$ 0.07  & 98.30 $\pm$ 0.04 \\
     2-layer Gravity AE & \cellcolor[gray]{.90} 75.84 $\pm$ 0.42 & \cellcolor[gray]{.90} 83.03 $\pm$ 0.22 & 97.77 $\pm$ 0.10  & 98.43 $\pm$ 0.10 \\
     3-layer Gravity AE & \cellcolor[gray]{.90} 74.61 $\pm$ 0.30 & \cellcolor[gray]{.90} 81.68 $\pm$ 0.26 & 97.58 $\pm$ 0.12  & 98.28 $\pm$ 0.11 \\
    \midrule   
    Linear Gravity VAE (ours) & \cellcolor[gray]{.90} 79.68 $\pm$ 0.36 & \cellcolor[gray]{.90} 85.00 $\pm$ 0.21 & 97.32 $\pm$ 0.06  & \cellcolor[gray]{.90} 98.26 $\pm$ 0.05 \\
     2-layer Gravity VAE  & \cellcolor[gray]{.90} 77.30 $\pm$ 0.81 & \cellcolor[gray]{.90} 82.64 $\pm$ 0.27 & 97.84 $\pm$ 0.25  & \cellcolor[gray]{.90} 98.18 $\pm$ 0.14 \\
     3-layer Gravity VAE  & \cellcolor[gray]{.90} 76.52 $\pm$ 0.61 & \cellcolor[gray]{.90} 80.73 $\pm$ 0.63 & 97.32 $\pm$ 0.23 & \cellcolor[gray]{.90} 97.81 $\pm$ 0.20 \\
    \bottomrule
  \end{tabular}
  \end{scriptsize}
\end{table}

\subsubsection{When (not) to use multi-layer GCN encoders?}

Linear graph AE and VAE reach strong empirical results on all graphs, and rival or outperform GCN-based graph AE and VAE in a majority of experiments. These models are also significantly simpler and more interpretable, each element of $z_i$ being interpreted as a weighted average from node $i$'s direct neighborhood. Therefore, we recommend the systematic use of linear graph AE and VAE as a first baseline, before diving into more complex encoding schemes whose actual benefit might be unclear.

Moreover, from our experiments, we also conjecture that multi-layer GCN encoders \textit{can} bring an empirical advantage when dealing with graphs with \textit{intrinsic non-trivial high-order interactions}. Notable examples of such graphs include the Amazon co-purchase graph (+5.61 AUC points for 2-layer GCN VAE) and web graphs such as Blogs, Google and Stanford, in which two-hop hyperlinks connections of pages usually include relevant information on the global network structure. On such graphs, capturing this additional information tends to improve results, especially 1) for the probabilistic VAE framework, and 2) when evaluating embeddings via the node clustering task (20+ AMI points on Blogs for 2-layer GCN AE/VAE) which is, by design, a more \textit{global} learning task than the quite \textit{local} link prediction problem. On the contrary, in citation graphs, the relevance of two-hop links is limited. Indeed, if a reference A in an article B cited by some authors is relevant to their work, authors will likely also cite this reference A, thus creating a one-hop link. Last, while the impact of the graph \textit{size} is unclear in our experiments (linear models achieve strong results even on large graphs, such as Patent), we note that graphs where multi-layer GCN encoders tend to outperform linear models are all relatively \textit{dense}.

To conclude, we conjecture that denser graphs with intrinsic high-order interactions (e.g. web graphs) should be better suited than the sparse Cora, Citeseer and Pubmed citation networks, to evaluate and to compare complex graph AE and VAE models, especially on global tasks such as node clustering.

\subsubsection{On $k$-hop linear encoders.} While, in this work, we only learn from direct neighbors interactions, variants of our models could capture higher-order links by considering polynomials of the matrix $A$. For instance, we could learn embeddings from one-hop and two-hop links by replacing $\tilde{A}$ by the normalized version of $A + \alpha A^2$ (with $\alpha > 0$), or simply $A^2$, in the linear encoders of Section 3.

Our online implementation proposes such alternative. We observed few to no improvement on most of our graphs, consistently with our claim on the effectiveness of simple one-hop strategies. Such variants also tend to increase running times (see below), as $A^2$ is usually denser than $A$.

\subsubsection{On running times.}
While this work put the emphasis on performance and not on training speed, we also note that linear AE and VAE models are 10\% to 15\% faster than their GCN-based counterparts. For instance, on an NVIDIA GTX 1080 GPU, we report a 6.03 seconds (vs 6.73 seconds) mean running time for training our linear graph VAE (vs 2-layer GCN graph VAE) on the featureless Citeseer dataset, and 791 seconds (vs 880 seconds) on the Patent dataset, using our sampling strategy from Section 4.1. This gain comes from the slightly fewer parameters and matrix operations required by one-hop linear encoders and from the sparsity of the one-hop matrix $\tilde{A}$ for most real-world graphs. 

Nonetheless, as an opening, we point out that the problem of scalable graph autoencoders remains quite open. Despite advances on the encoder, the standard inner-product decoder still suffer from a $O(d n^2)$ time complexity. Our very simple sampling strategy to overcome this quadratic cost on large graphs (randomly sampling subgraphs to reconstruct) might not be optimal. Future works will therefore tackle these issues, aiming at providing more efficient strategies to scale graph AE and VAE to large graphs with millions of nodes and edges.

\section{Conclusion}

Graph autoencoders (AE), graph variational autoencoders (VAE) and most of their extensions rely on multi-layer graph convolutional networks (GCN) encoders to learn node embedding representations. In this paper, we highlighted that, despite their prevalent use, these encoders are often unnecessarily complex. In this direction, we introduced and studied significantly simpler versions of these models, leveraging one-hop linear encoding strategies. Using these alternative models, we reached competitive empirical performances w.r.t. GCN-based graph AE and VAE on numerous real-world graphs. We identified the settings where simple one-hop linear encoders appear as an effective alternative to multi-layer GCNs, and as first relevant baseline to implement before diving into more complex models. We also questioned the relevance of repeatedly using the same sparse medium-size datasets (Cora, Citeseer, Pubmed) to evaluate and to compare complex graph AE and VAE models. 

\bibliographystyle{splncs04}
\bibliography{references}

\end{document}